\documentclass[a4paper,conference]{IEEEtran}

\usepackage{ifpdf}
\usepackage{cite}
\usepackage[pdftex]{graphicx}
\usepackage{array}
\usepackage{mdwmath}
\usepackage{mdwtab}
\usepackage{amssymb,latexsym}
\usepackage{stfloats}
\usepackage{amsmath}
\usepackage{subfig}
\usepackage{todonotes}

\usepackage{hyperref}
\hypersetup{
    colorlinks=true,
}

\usepackage[numbers]{natbib}
\usepackage{algcompatible}

\usepackage{algorithm}

\usepackage[font={footnotesize}]{caption}

\usepackage{balance}

\usepackage[utf8]{inputenc}

\usepackage[T1]{fontenc}

\usepackage[compatible]{algpseudocode}

\renewcommand{\figurename}{Figure}

\usepackage[most]{tcolorbox}
\newtcolorbox{highlighted}{colback=yellow,breakable}

\DeclareRobustCommand*{\IEEEauthorrefmark}[1]{%
\raisebox{0pt}[0pt][0pt]{\textsuperscript{\footnotesize\ensuremath{#1}}}}

\renewcommand\IEEEkeywordsname{Keywords}

\makeatletter
\def\ps@IEEEtitlepagestyle{%
  \def\@oddfoot{\mycopyrightnotice}%
  \def\@evenfoot{}%
}
\def\mycopyrightnotice{%
  {\footnotesize 979-8-3315-9679-8/25/\$31.00 ©2025 IEEE\hfill}
  \gdef\mycopyrightnotice{}
}

\begin{document}

\title{Skin Color Measurement from Dermatoscopic Images: An Evaluation on a Synthetic Dataset} 
\author{\IEEEauthorblockN{
Marin Benčević\IEEEauthorrefmark{1},
Robert Šojo\IEEEauthorrefmark{1},
Irena Galić\IEEEauthorrefmark{1}}
\IEEEauthorblockA{\IEEEauthorrefmark{1}
Faculty of Electrical Engineering, Computer Science and Information Technology, Osijek, Croatia}
{\it marin.bencevic@ferit.hr}}

\maketitle

\begin{abstract}
This paper presents a comprehensive evaluation of skin color measurement methods from dermatoscopic images using a synthetic dataset (S-SYNTH) with controlled ground-truth melanin content, lesion shapes, hair models, and 18 distinct lighting conditions. This allows for rigorous assessment of the robustness and invariance to lighting conditions. We assess four classes of image colorimetry approaches: segmentation-based, patch-based, color quantization, and neural networks. We use these methods to estimate the Individual Typology Angle (ITA) and Fitzpatrick types from dermatoscopic images. Our results show that segmentation-based and color quantization methods yield robust, lighting-invariant estimates, whereas patch-based approaches exhibit significant lighting-dependent biases that require calibration. Furthermore, neural network models, particularly when combined with heavy blurring to reduce overfitting, can provide light-invariant Fitzpatrick predictions, although their generalization to real-world images remains unverified. We conclude with practical recommendations for designing fair and reliable skin color estimation methods.
\end{abstract}

\begin{IEEEkeywords}
AI fairness; biometrics; colorimetry; dermatoscopy
\end{IEEEkeywords}

\IEEEpeerreviewmaketitle

\section{Introduction}
\label{sec1}

Deep neural networks for classifying dermatoscopic images are widely used and are a very active field of research \cite{kurtanskyEffectPatientContextual2024, estevaDermatologistlevelClassificationSkin2017}. However, studies on clinical images have shown that neural networks can have marked biases against dark-skinned individuals \cite{daneshjouDisparitiesDermatologyAI2022, bevanDetectingMelanomaFairly2022}. To see if the same sort of bias also extends to dermatoscopic images, researchers must group the subjects by skin color. Ideally, this would be done with separate skin color measurements for each subject; however, commonly used dermatological datasets do not include such measurements. Instead, researchers often estimate skin color directly from the image pixels, relying on image colorimetry methods. These estimates can be sensitive to lighting, image composition, and variations in acquisition protocols. This paper aims to shed light on how lighting conditions can impact skin color estimates from image pixels, as well as how different methods compare to each other.

Building on the insights by \citet{kalbRevisitingSkinTone2023}, the present work aims to systematically evaluate colorimetry-based skin color measurement techniques. Our central contribution is the use of a publicly available system for rendering dermatoscopic images called S-SYNTH \cite{kimSSYNTHKnowledgeBasedSynthetic2024} in which the \textit{ground truth} (GT) melanin content of the skin can be precisely specified during rendering. This allows us to experimentally vary the lighting conditions while keeping all other variables constant.

We compare three colorimetric techniques and a neural network classifier: 
\textbf{(1) segmentation-based ITA estimation} is the most consistent but demands reliable segmentation; \textbf{(2) color quantization} achieves comparable 
performance without segmentation; and \textbf{(3) patch-based approaches} show substantial, lighting-dependent biases. Meanwhile, a \textbf{(4) neural 
network} trained via ordinal regression outperforms other methods for Fitzpatrick type classification, at the cost of requiring labeled training data and unknown generalization performance. All code and data is available at \href{https://github.com/marinbenc/dermatoscopy_colorimetry_eval}{github.com/marinbenc/dermatoscopy\_colorimetry\_eval}.

\section{Background and Related Work}

\citet{kalbRevisitingSkinTone2023} compared four methods of estimating skin colors from dermatoscopic images approaches and identified substantial discrepancies among them, underscoring the potential for erroneous or inconsistent measurements. We extend their work by including other classes of colorimetry methods as well as evaluating the methods on a synthetic dataset with known ground-truth melanin content.

To do so, we adopt two widely used metrics for assessing skin tone: the Fitzpatrick scale (FP) \cite{fitzpatrickValidityPracticalitySunReactive1988} and the Individual Typology Angle (ITA) \cite{farkasInternationalAnthropometricStudy2005}, both frequently employed in skin color bias research \cite{grohEvaluatingDeepNeural2021, kinyanjuiEstimatingSkinTone2019}.

\subsubsection{Fitzpatrick scale} 

This categorization divides skin types into six groups based on UV response, from Type~I (very fair, always burns) to Type~VI (deeply pigmented, never burns). Although labeling Fitzpatrick types from images can be subjective \cite{grohEvaluatingDeepNeural2021}, it serves as a useful framework for large-scale bias analyses.

\subsubsection{Individual Typology Angle}

ITA provides a continuous measure of constitutive pigmentation, with higher values corresponding to lighter skin. Image pixels are transformed into CIE-Lab space and the following equation is applied \cite{kinyanjuiEstimatingSkinTone2019}:
\begin{equation}\label{eq:ita}
  \mathrm{ITA}(L^*, b^*) = \arctan\!\Bigl(\frac{L^* - 50}{b^*}\Bigr)\,\frac{180}{\pi},
\end{equation}
where \(L^*\) denotes lightness and \(b^*\) the blue--yellow axis.

\section{Methods}

The synthetic dataset comprises 10,000 dermatoscopic images and segmentation masks generated by varying melanin content, lesion shape, hair models, and 18 distinct lighting conditions, among other factors, following the procedure outlined in \cite{kimSSYNTHKnowledgeBasedSynthetic2024}. The dataset is well-balanced across each of the image generation parameters.

\subsection{Obtaining the Ground-Truth Fitzpatrick Type Labels}

To obtain ground-truth Fitzpatrick (FP) labels, we use the segmentation masks provided by S-SYNTH \cite{kimSSYNTHKnowledgeBasedSynthetic2024} to remove lesion regions and compute the mean ITA value for each image, discarding outlier pixels. The resulting ITA values are then binned into FP types using the thresholds from \cite{kinyanjuiEstimatingSkinTone2019}. We empirically derive thresholds for binning by computing the mean melanosome fraction within each FP bin and taking the mid-point between each mean as the threshold values for binning.

By defining the synthetic Fitzpatrick labels directly from the melanosome fraction rather than from pixel-based color values, we ensure that these labels remain invariant to the lighting variations in the dataset.

\subsection{Dermatoscopy Colorimetry Methods}
\label{sec:colorimetry}

In this study, we evaluate a representative implementation of four general classes of skin color estimation methods from dermatoscopic images:
(\emph{1}) \textbf{segmentation-based}, 
(\emph{2}) \textbf{patch-based}, 
(\emph{3}) \textbf{color quantization}, and 
(\emph{4}) \textbf{model-based} approaches. 
Below, we briefly describe each category, provide related work, and outline the specific implementation details used in our experiments.

\subsubsection{Segmentation-based}
In this class of methods, healthy skin regions are isolated by removing lesions (and sometimes other artifacts). 
Traditional approaches employ image processing or deep neural networks for segmentation, then compute the mean color 
(e.g., in CIELAB space) of the remaining pixels, discarding outliers \cite{kinyanjuiEstimatingSkinTone2019}. Thus, any segmentation biases can propagate to the color estimation. 
In our work, we use ground-truth segmentation masks to evaluate the best-case scenario of segmentation-based approaches. We discard lesion pixels, compute the median \(L^*\) and \(b^*\) values of the remaining skin, excluding values lying 
more than one standard deviation away from the medians, and then calculate the ITA (Eq.~\ref{eq:ita}) from the remaining median values without outliers.

\subsubsection{Patch-based}
Instead of segmenting lesions, \citet{bevanDetectingMelanomaFairly2022} propose sampling small patches around the image 
edges, then choosing the ``lightest'' patch (highest ITA) under the assumption that normal skin regions are brighter than 
lesions, hairs and other artefacts. Note that this assumption does not account for hypopigmentation or lighter hairs, both common in dark-skinned individuals. Following \cite{kalbRevisitingSkinTone2023}, we reproduce the method from \cite{bevanDetectingMelanomaFairly2022} by replacing \(\arctan\) 
with \(\arctan2\) when computing ITA. We further improve the method by calibrating the resulting estimates to match 
segmentation-based ITA values, as the raw patch-based outputs show a consistent bias under certain lighting conditions (see Fig. \ref{fig:ita_bland_altman}).

\subsubsection{Color quantization}
Clustering-based methods sidestep lesion segmentation by quantizing the image to a small set of dominant colors, then 
selecting the color that most likely represents normal skin \cite{bencevicUnderstandingSkinColor2024a}. In our implementation, 
we first apply CLAHE on the \(L^*\) channel and use Otsu thresholding of the HSV value channel, followed by morphological 
expansion, to mask out lesions and other artifacts. We then convert the masked skin regions back to CIELAB and perform 
k-means clustering, with the optimal number of clusters determined as in \cite{satopaFindingKneedleHaystack2023}. 
We compute the ITA as in Eq.~\ref{eq:ita} from the most populated cluster.

\subsubsection{Model-based}
Finally, deep learning methods can be trained to directly predict skin color categories. Such models may learn invariances 
to illumination or image composition but require sufficient labeled data \cite{grohEvaluatingDeepNeural2021}. We implement 
an ordinal regression framework (CORAL \cite{coral2020}) on top of a VGG11 \cite{simonyanVeryDeepConvolutional2015} backbone, 
with Fitzpatrick type labels derived by binning melanosome fractions used during image generation (threshold values are available in the supplied codebase). 

The 
10{,}000 synthetic images are split into 80\% training, 10\% validation, and 10\% testing sets. Each image is downscaled to $224 \times 224$ pixels, and a heavy Gaussian 
blur a standard deviation of 5 and a kernel size of 21 is applied to each input before feeding it to the network to reduce overfitting to the synthetic image generation method. After 100 training epochs, we select the best model checkpoint based on validation loss. All reported results 
are from evaluating the model on the test set.

\section{Results}

\begin{figure}[b]
\centerline{\includegraphics[width=\columnwidth]{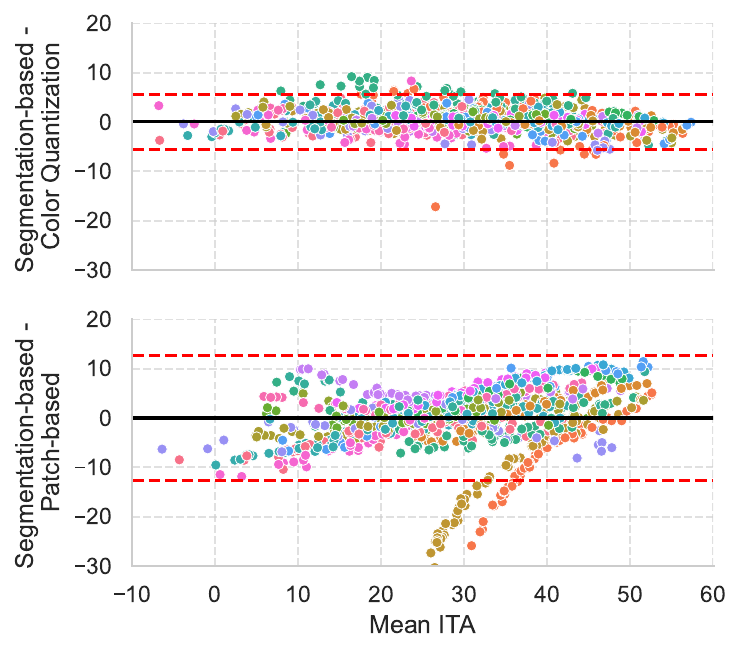}}
\caption{Bland-Altman plots comparing the color quantization and the patch-based to the segmentation-based ITA estimates. Colors indicate different lighting conditions used during image generation, a total of 18 categories.}
\label{fig:ita_bland_altman}
\end{figure}

We first compare the three ITA measurement approaches described in Section~\ref{sec:colorimetry}: 
segmentation-based, patch-based, and color quantization. The patch-based 
approach exhibited a consistent linear bias relative to the segmentation-based ITA values under most lighting conditions, prompting us to apply an ordinary least squares (OLS) regression to calibrate its estimates.

Our assumption is that a good estimate of ITA would be perfectly correlated with the melanosome fraction used during the synthetic image generation, as ITA should represent the melanin content of the normal skin region regardless of lighting conditions, hair, lesions or other artefacts. Thus, to measure the quality of each prediction method, we compute Pearson's correlation between each ITA estimate and the ground-truth melanosome fraction, 
and generate 95\% confidence intervals via 1{,}000 bootstrap resamples. Table~\ref{tab:ita_corr_ci} presents the results 
for the segmentation-based, color quantization, and calibrated patch-based methods.
\begin{table}[h!]
    \centering
    \caption{Pearson's correlation (\(\rho\)) between each ITA estimate and ground-truth melanosome fraction, with
    95\% confidence intervals (CIs) over 1{,}000 bootstrap resamples.}
    \label{tab:ita_corr_ci}
    \def\arraystretch{1.2}
    \begin{tabular}{lcc}
        \textbf{Method} & \boldmath$\rho$ & \textbf{95\% CI} \\
        Segmentation-based & 0.705 & [0.695, 0.714] \\
        Color quantization & 0.683 & [0.671, 0.693] \\
        Calibrated patch-based & 0.624 & [0.610, 0.638]
    \end{tabular}
\end{table}

A Bland--Altman analysis (Fig.~\ref{fig:ita_bland_altman}) reveals that the color quantization method agrees 
more closely with the segmentation-based ITA, whereas the patch-based method exhibits a consistent bias relative to 
segmentation. This discrepancy is further amplified under certain lighting conditions, even after calibration.

\begin{figure*}[htbp]
\centerline{\includegraphics[width=\textwidth]{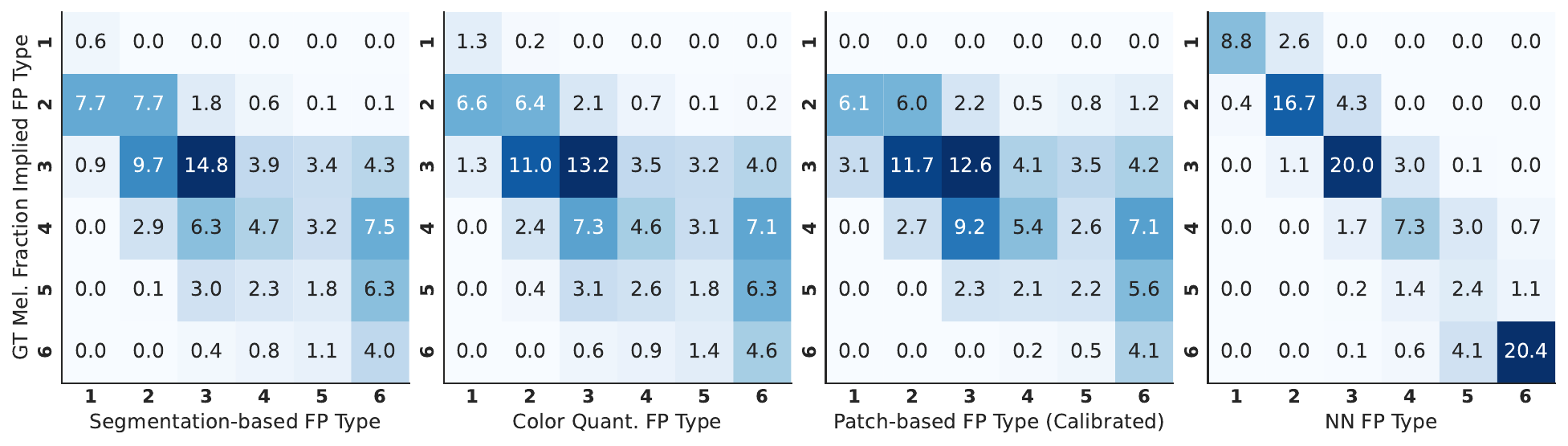}}
\caption{Confusion matrices of Fitzpatrick (FP) type estimates as compared to the ground-truth FP, calculated by binning the melanosome fraction. Numbers indicate \% the total number of test images ($n = 1000$).}
\label{fig:fp_confusion_matrices}
\end{figure*}

\subsection{Sensitivity to Lighting Conditions}

\noindent
ITA measurements should ideally capture melanin content rather than external illumination. To assess this, we fit two 
linear models for each estimation method (segmentation-based, color quantization, and calibrated patch-based): $\mathrm{ITA} \sim \mathrm{mel}$, and $\mathrm{ITA} \sim \mathrm{mel} + \mathrm{light}$,
where  \(\mathrm{mel}\) is the melanosome fraction used during synthetic image generation and \(\mathrm{light}\) is a categorical variable indicating the 18 lighting conditions. We then compute the coefficient 
of determination (\(R^2\)) for both models and measure how much adding \(\mathrm{light}\) improves the fit 
(\(\Delta R^2\)). Table~\ref{tab:lighting_invariance} summarizes these results.

\begin{table}[h!]
    \centering
    \caption{Coefficient of determination $R^2$ under two linear models of ITA, with and without including lighting conditions. \(\Delta R^2\) indicates the difference between the two $R^2$ values, i.e. the relative sensitivity of each ITA estimate method to lighting conditions.}
    \label{tab:lighting_invariance}
    \begin{tabular}{lccc}
        \textbf{Method} & $ITA \sim mel$ & $ITA \sim mel + light$ & $\Delta R^2$ \\
        Segmentation-based    & 0.5 & 0.72 & 0.23 \\
        Color quantization    & 0.47 & 0.678 & 0.21 \\
        Calibrated patch-based & 0.39 & 0.7 & 0.31 \\
    \end{tabular}
\end{table}

\noindent
From Table~\ref{tab:lighting_invariance}, adding the lighting term increases \(R^2\) for each method, revealing that lighting still exerts a significant influence on ITA estimates. Notably, the calibrated patch-based approach shows the largest \(\Delta R^2\) (0.306), indicating a higher sensitivity to illumination compared to the other methods.

\subsection{Fitzpatrick Type Estimation}

We evaluate the ability of each method to predict the Fitzpatrick type (FP) of an image, using FP labels based on empirically binning the melanosome fraction values. For the three ITA-based approaches (segmentation-based, 
patch-based, and color quantization), we convert the estimated ITA values into FP types using the commonly-used thresholds proposed 
in \cite{kinyanjuiEstimatingSkinTone2019}.

Fig.~\ref{fig:fp_confusion_matrices} shows the confusion matrices for each of the four methods compared against the 
ground-truth FP labels. Table~\ref{tab:fp_estimation_results} summarizes the balanced accuracy, precision, recall, F1 score, 
and quadratically weighted kappa for each approach.

\begin{table}[h!]
    \centering
    \caption{Fitzpatrick Type Estimation Results: Balanced Accuracy (Acc), Precision (Prec), Recall (Rec), 
    F1 score (F1), and Quadratically Weighted Kappa (Kappa).}
    \label{tab:fp_estimation_results}
    \begin{tabular}{lccccc}
    \textbf{Method} & \textbf{Acc} & \textbf{Prec} & \textbf{Rec} & \textbf{F1} & \textbf{Kappa} \\
    Segmentation-based & 0.29 & 0.46 & 0.29 & 0.28 & 0.59 \\
    Color quantization & 0.29 & 0.43 & 0.29 & 0.29 & 0.59 \\
    Calibrated patch-based & 0.27 & 0.32 & 0.27 & 0.25 & 0.52 \\
    Model-based & 0.72 & 0.71 & 0.72 & 0.70 & 0.95 \\
    \end{tabular}
\end{table}

\noindent
From Table~\ref{tab:fp_estimation_results}, the model-based approach achieves the highest performance on every 
metric. Among the ITA-based methods, segmentation-based and 
color quantization exhibit slightly higher balanced accuracy and kappa than the patch-based approach. The confusion matrices (Fig.~\ref{fig:fp_confusion_matrices}) reveal that errors in the ITA-based 
methods often involve adjacent FP classes.

\section{Conclusion and Discussion}

Our results indicate that segmentation-based ITA estimation yields the most robust performance across varied lighting conditions, 
but this method inherently depends on accurate lesion segmentation---itself potentially prone to bias and error when derived from 
learned models. By contrast, color quantization avoids the need for segmentation and achieves comparable accuracy and lighting 
invariance, making it a compelling, interpretable alternative that relies solely on simple clustering rather than complex models.

Patch-based methods exhibit substantial bias tied to individual lighting setups. In practice, each lighting condition 
appears to introduce a consistent offset that would necessitate calibration if the approach were to be used reliably. Authors proposing new colorimetry 
techniques should account for these pitfalls and consider calibration strategies for different lighting scenarios.

Our experiments show that neural networks, when combined with heavy blurring to mitigate overfitting on the synthetic dataset, can yield robust light-invariant Fitzpatrick type predictions. However, without large, well-labeled real-world datasets, their generalizability remains unverified, indicating a promising direction for future research.

In summary, we recommend that researchers favor either color quantization or segmentation-based approaches for skin color estimation. These methods have proven more robust and less prone to bias than heuristic strategies, such as selecting the lightest patch, which can systematically introduce errors under varying lighting conditions. We further advise that any colorimetric measurements be calibrated against multiple reference measures and datasets to account for the fixed biases induced by illumination. Finally, neural network approaches show promise for obtaining light-invariant predictions, representing a promising avenue for future research.

\section*{Acknowledgment}

This work has been supported by the Faculty of Electrical Engineering, Computer Science and Information Technology Osijek grant ``IZIP 2024''. The authors would like to thank \citet{kimSSYNTHKnowledgeBasedSynthetic2024} for the use of their image generation framework.

\bibliographystyle{ieeetr}
\bibliography{bibliography}

\end{document}